\documentclass[letterpaper]{article} 
\usepackage{aaai23}  
\usepackage{times}  
\usepackage{helvet}  
\usepackage{courier}  
\usepackage[hyphens]{url}  
\usepackage{graphicx} 
\usepackage{makecell}
\usepackage{amsmath}
\usepackage{natbib}  
\usepackage{caption} 
\usepackage{algorithm}
\usepackage{algorithmic}

\usepackage{newfloat}
\usepackage{listings}
\DeclareCaptionStyle{ruled}{labelfont=normalfont,labelsep=colon,strut=off} 
\floatstyle{ruled}
\newfloat{listing}{tb}{lst}{}
\floatname{listing}{Listing}
\title{BIFRNet: A Brain-Inspired Feature Restoration DNN for Partially Occluded Image Recognition}
\author{
    Jiahong Zhang \textsuperscript{\rm 1,2},
    Lihong Cao \textsuperscript{\rm 1,2 
    \thanks{Corresponding author}
    }, 
    Qiuxia Lai \textsuperscript{\rm 1,2},
    Bingyao Li \textsuperscript{\rm 2},
    Yunxiao Qin \textsuperscript{\rm 1,2} \\
}
\affiliations{
    \textsuperscript{\rm 1} State Key Laboratory of Media Convergence and
Communication, Communication University of China, Beijing, China \\
    \textsuperscript{\rm 2} Neuroscience and Intelligent Media Institute, Communication University of China, Beijing, China \\
    \{zhangjh, lihong.cao, qinyunxiao, qxlai\}@cuc.edu.cn, l\_bingyao@163.com
}

\usepackage{bibentry}

\begin{document}

\maketitle

\begin{abstract}
The partially occluded image recognition (POIR) problem has been a challenge for artificial intelligence for a long time.
A common strategy to handle the POIR problem is using the non-occluded features for classification. 
Unfortunately, this strategy will lose effectiveness when the image is severely occluded, since the visible parts can only provide limited information. 
Several studies in neuroscience reveal that feature restoration which fills in the occluded information and is called amodal completion is essential for human brains to recognize partially occluded images. 
However, feature restoration is commonly ignored by CNNs, which may be the reason why CNNs are ineffective for the POIR problem.
Inspired by this, we propose a novel brain-inspired feature restoration network (BIFRNet) to solve the POIR problem.
It mimics a ventral visual pathway to extract image features and a dorsal visual pathway to distinguish occluded and visible image regions.
In addition, it also uses a knowledge module to store object prior knowledge and uses a completion module to restore occluded features based on visible features and prior knowledge.
Thorough experiments on synthetic and real-world occluded image datasets show that BIFRNet outperforms the existing methods in solving the POIR problem. Especially for severely occluded images, BIRFRNet surpasses other methods by a large margin and is close to the human brain performance. Furthermore, the brain-inspired design makes BIFRNet more interpretable. 
\end{abstract}

\section{Introduction}\label{sec1}
Convolutional neural networks (CNNs) have achieved remarkable results in computer vision and even reached human-level performance in some tasks such as face recognition \cite{face}. 
However, the robustness of CNNs under partial occlusion is much lower than that of the human visual system \cite{geirhos2017comparing}, possibly because occluders can easily mislead CNNs to predict wrong results. 
Since occlusion occurs in many real-world images or videos, most computer vision tasks, such as human-computer interaction \cite{liu2020fpha}, instance segmentation \cite{ke2021deep} and human pose estimation \cite{das2020end} are troubled by the partial occlusion problem.
In this paper, we try to solve the partially occluded image recognition (POIR) problem.


A natural idea to solve the POIR problem is extracting robust features. For example, feature regularization \cite{liao2016learning} and convolution kernel regularization \cite{tabernik2016towards} were used to force feature activations to be disentangled for different objects. 
However, the paper \cite{Kortylewski_2020_WACV} indicated that these methods are not always robust to partial occlusion. Some works proposed that using generative models to restore the clean image from the partially occluded image before classification usually improves the classification accuracy \cite{2016Plug,2018Generative,Yuan_2019_ICCV,SDGAN}. 
Nevertheless, image-level restoration commonly results in costly computation, and training a generative model is challenging. 

A recent popular direction to solve the POIR problem is removing occluded features (ROF). 
It removes the features that are corrupted by occluders and then utilizes the remaining visible features for classification. For example, an external sub-network can be used to generate mask-guided attention which discriminates occluded and visible features \cite{maskatt,zhang2018occluded,mask2,mask,mask4,chen2021occlude,mask3}. Networks then classify images based only on the visible features. 

Although ROF is effective to solve the POIR problem, it does not restore occluded features. 
In contrast, several works in neuroscience demonstrated that amodal completion which restores features in human brain is important for humans to recognize partially occluded objects \cite{amodalv4}. 
Amodal completion extends edges behind occluders if a continuous smooth connection exists \cite{kellman1991theory} and fills missing surfaces \cite{he1992surfaces} and volumes \cite{tse1999volume}.  In addition, the work \cite{peters2021capturing} indicated that amodal completion relies on
prior knowledge about image statistics.

Based on this, we suggest that an occlusion robustness model should be able to restore the occluded features.
To obtain this ability, the model should firstly be able to
distinguish visible and occluded parts and should have prior knowledge of the target objects.
\begin{figure*}[th]%
	\centering
	\includegraphics[width=1\textwidth]{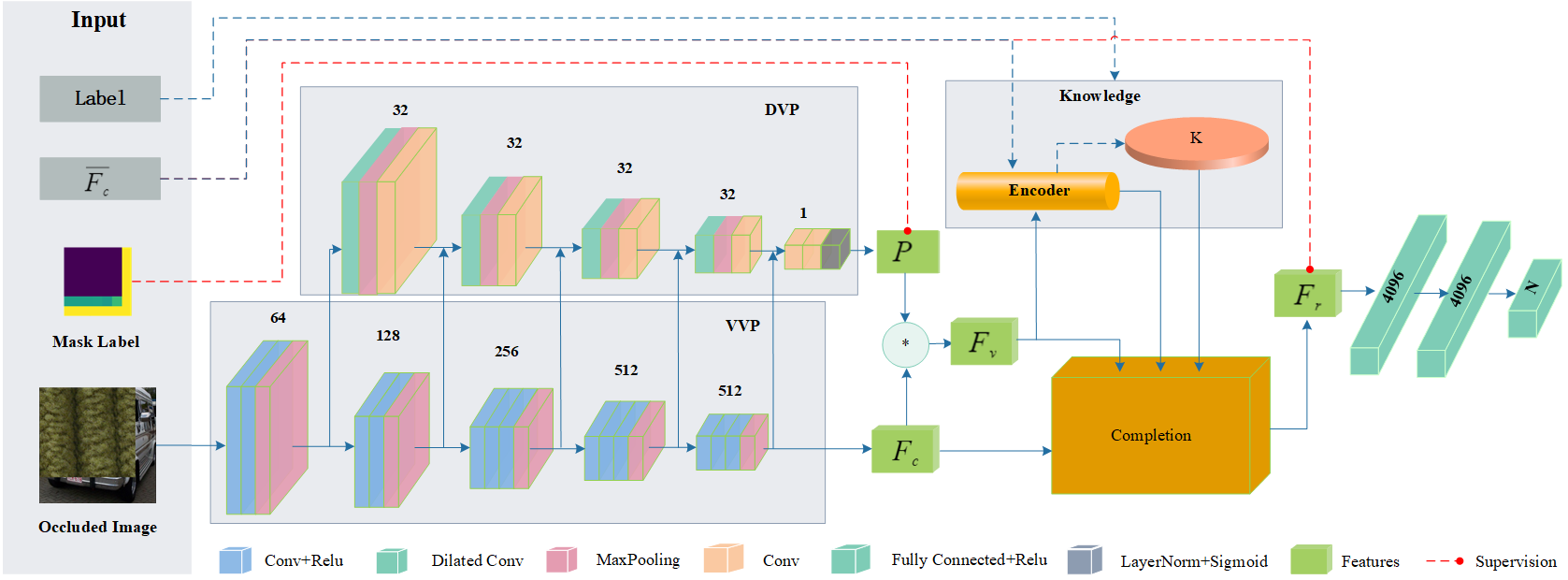}
	\caption{The architecture of BIFRNet. The ventral visual pathway (VVP) is based on VGG16, extracting image features $F_c$. The dorsal visual pathway (DVP) which consists of four D+MP+C blocks and a C+C+LayerNorm+Sigmoid block outputs spatial attention $P$. $F_v$ is the visible feature obtained by the Hadamard product of $F_c$ and $P$. The knowledge module forms prior knowledge $\mathcal K$ by training. $\mathcal K$ is input into the completion module, where restored features $F_r$ is generated for classification. $\bar{F_c}$ is the clear image feature from the pre-trained VGG16 pool5 layer and is used as the label of $F_r$. The dotted line connections exist only in the training phase and do not exist in the testing phase. Red dotted lines denote the paths where we compute loss. Numbers in this figure denote the feature channels. $\bar{F_c}$ and $F_v$ are the inputs of the encoder in training and testing, respectively. 
	}
	\label{architecture}
\end{figure*}

We introduce a novel brain-inspired feature restoration network (BIFRNet) with all these capabilities, as shown in Figure~\ref{architecture}. Inspired by human visual system, we construct a ventral visual pathway (VVP) to extract image features and a dorsal visual pathway (DVP) to generate spatial attention for distinguishing occlusion \cite{roleofattention,humanatt,humanatt2}.
Amodal completion can be construed as an inference process which restores the occluded features based not only on the visible features but also prior knowledge about target objects \cite{peters2021capturing}. Therefore, a completion module is proposed to imitate this process and a knowledge module is proposed to provide the prior knowledge. Restoration here is feature-level and differs from the existing image-level restoration method \cite{2016Plug,2018Generative,SDGAN} and is close to human brain.
We review relevant neuroscience bases in Related Work and carry out ablation experiments of BIFRNet in Discussion, showing its interpretability and effectiveness. 
In conclusion, the main contributions of this paper include:

1) We propose a brain-inspired feature restoration network, BIFRNet. It mimics human ventral visual pathway to extract features and dorsal visual pathway to discriminate visible and occluded parts. It also has a knowledge module to provide prior knowledge and a completion module to restore the occluded features. 

2) Studies of BIFRNet's components show their designated functions and effectiveness. The brain-inspired design provides BIFRNet with high interpretability and may benefit future research for partially occluded image recognition-like problems. 

3) BIFRNet gets state-of-the-art performance on synthetic and real-world occluded images. On the synthetic dataset, it achieves human-like performance and outperforms the state-of-the-art model by $7\%$ recognition accuracy for highly occluded images.

\section{Related Work}\label{s2}
\subsection{CNNs for Occluded Image Recognition}
Although CNN models are successful in image recognition, they suffer from the POIR problem. Data augmentation in terms of partial occlusion \cite{cutout,cutmix} is a common method to solve this problem. However, the work  \cite{Kortylewski_2020_WACV} indicated that it is not always robust. 
Recently, removing occluded features (ROF)-based methods become popular for POIR. It removes the features corrupted by occluders and then utilizes the remaining visible features for classification. 
Common ROF-based methods include attention-based and compositional models. There are many variants of attention mechanisms in dealing with occlusion \cite{zhang2018occluded,topdown}. Top-down feedback attention in TDAPNet is composed of explainable part attention, which can reduce the contamination of occlusion \cite{topdown}. 
Mask-guided attention is generated by an extra sub-network to distinguish occluded and non-occluded features. 
For example, MaskNet\cite{maskatt} used mask attention to assign higher weights to the hidden units associated to the visible facial parts and achieves occlusion robustness in face recognition. 
Pairwise Differential Siamese Network (PDSN) \cite{maskdictonary} learned the correspondence between occluded facial regions and corrupted activations, and used the visible features to classify. 
Mask-guided attention was also used in other vision tasks such as occluded pedestrian detection \cite{mask4,chen2021occlude}. 
Compositional models achieved interpretable occlusion robustness by combining CNN and probability models \cite{wang2017detecting,zhang2018deepvoting,Kortylewski_2020_WACV}.
Kortylewski et al. \cite{kortylewski2020combining} proposed a dictionary-based compositional model to learn the feature dictionary for better classification. They also presented a differentiable compositional model that can localize occluders and classify images by the visible features \cite{Kortylewski_2020_WACV}. However, compositional models are complex and require manually designed constraints. 

\subsection{Neuroscience Basis for Occluded Image Recognition} \label{humanbase}
It is generally believed that the human visual system includes ventral and dorsal visual pathways \cite{twostream}. The ventral pathway works as an object recognizer for objects at the center view. The dorsal pathway is to recognize objects' spatial positions and movements. When an object is partially occluded, the human visual system may have two strategies to restore the missing information: modal completion and amodal completion \cite{michotte1991amodal}. Modal completion is a process of producing an illusory contour effect \cite{kanizsa1979organization}. Amodal completion is the ability to see an entire object despite parts of it being covered by another object in front of it, which is more suitable for solving the POIR problem than modal completion \cite{johnson2005recognition}. Studies have shown that the completion of simple lines and shapes occurs early in the ventral visual pathway \cite{lee2003computations,1111}, and simple CNN networks such as AlexNet also have similar capabilities \cite{kim2021neural}. However, complex completion needs to judge the relationship between the occluded object and the occluder \cite{johnson2005recognition}. This completion requires more time and depends on the recurrent process \cite{sekuler1992perception,shore1997shape}. In this paper, we model the relationship between the occluded object and the occluder by spatial attention and use the recurrent neural network to model the completion function. In addition, amodal completion must rely on prior knowledge about the statistics of images or about the shape of objects \cite{peters2021capturing}.

\section{Proposed Method}\label{s3}
\begin{figure}[h]%
	\centering
	\includegraphics[width=\linewidth]{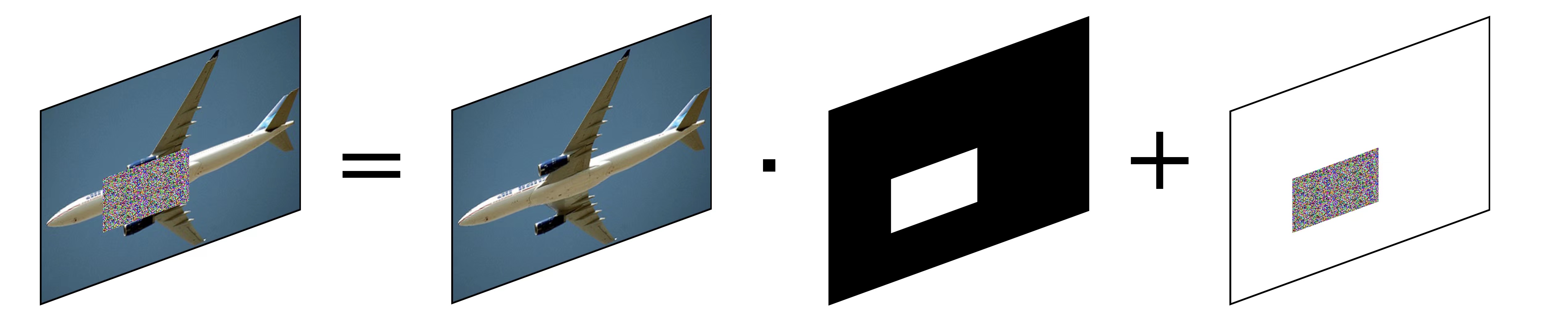}
	\caption{The visualization of Equation \ref{e2}. Here, black elements in the matrix are 1, and white elements are 0.}
	\label{occget}
\end{figure}

%
The prediction of a normal CNN classification model can be fomulated as
\begin{equation} 
	p = f(x),
\end{equation}
where $x$ is an image, $f$ is the CNN model, and $p$ is the predicted classification probability. 
In occlusion scene, an occluded image can be represented as:
\begin{equation}
	x_{occ} = x \cdot o + m,
	\label{e2}
\end{equation}
where $o$ is a binary occlusion position matrix with the same size as $x$. 0 and 1 in $o$ denote the positions for occluded and non-occluded, respectively. $m$ is the occluder, and $\cdot$ refers to element-wise production. 
Figure~\ref{occget} visualizes this equation.
Disturbed by the occluder, CNN models commonly misclassify $x_{occ}$, which is known as the POIR problem. This paper proposes BIFRNet to solve the POIR problem.

\subsection{Network Architecture} \label{framwork}
Figure~\ref{architecture} shows the proposed BIFRNet and Table~\ref{featured} lists the descriptions of abbreviations. BIFRNet uses three main components to achieve occlusion robustness: 1) the visual pathways VVP and DVP, 2) the knowledge module, and 3) the completion module. VVP extracts image features and DVP generates the spatial attention for distinguishing occlusion. During training, prior knowledge of target objects will be formed in the knowledge module. The completion module restores occluded features with visible features and the knowledge. Finally, several fully connected layers are used for classification based on the restored features. The input of BIFRNet is an image with the size of $C\times H\times W$, and the features $F_c$, $F_v$ and $F_r$ are with size of $C_v \times H_v \times W_v$.

\begin{table}  
	\renewcommand\arraystretch{1.5}
	\centering
	\begin{tabular}{ll} 
		\hline                      
		\textbf{Feature} & Description  \\  
		\hline  
		$F_c$  & \makecell[l]{The features of the ventral visual pathway} \\
		$\bar{F_c}$  & \makecell[l]{The clear image features from the pre-trained \\ VGG16 pool5 layer, used as the label of $F_r$} \\
		$P$ & \makecell[l]{The spatial attention from the dorsal visual \\ pathway} \\
		$F_v$  & \makecell[l]{The visible features obtained by multiplying \\ $F_c$ by $P$}\\ 
		$F_r$  & The restored features 	\\  
		$F_k$  & The knowledge coding 	\\  
		$\mathcal K$ & The knowledge matrix \\
		\hline 
	\end{tabular}  
 \caption{Descriptions of the feature symbols in BIFRNet}
	\label{featured}
 \end{table}

\begin{figure*}[t]%
	\centering
 \includegraphics[width=0.85\textwidth]{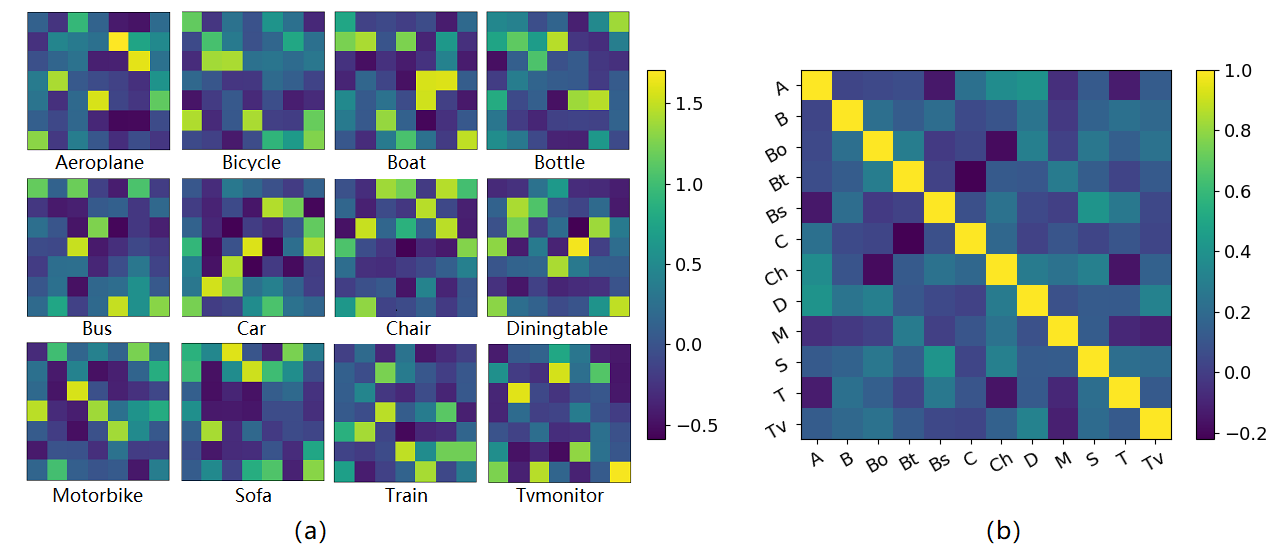}
	\caption{Illustration of $\mathcal K$ after training on the synthetic Occluded-Vehicles dataset \cite{Kortylewski_2020_WACV}. Figures in (a) are the knowledge representations of 12 categories, aeroplane, bicycle, boat, bottle, bus, car, chair, diningtable, motorbike, sofa, train and tvmonitor in the dataset. (b) shows the similarity matrix of different category knowledge in $\mathcal K$, the names of which are abbreviated to A, B, Bo, Bt, Bs, C, Ch, D, M, S, T and Tv in sequence. Cosine similarity is used as the similarity measure. It is shown that $\mathcal K$ has well separable representations of the 12 categories.}
	\label{fusionm}
\end{figure*}

\subsubsection{Visual Pathways}
Visual pathways of BIFRNet consist of VVP and DVP. VVP is VGG16 \cite{VGG16} without fully connection layers. DVP is parallel to VVP, which consists of 4 D+MP+C and one C+C+LayerNorm+Sigmoid blocks in series, where D is dilated convolution layer with dilated rate 2, C is convolution layer, LayerNorm is layer normalization \cite{ln}, and MP is Maxpooling layer. Dilated convolution is used to increase the receptive field and form good attention. 
Before every block in DVP, there are connections with VVP, where the features are concatenated. These connections allow DVP to consider the information from multiple layers in VVP. 
To get a trade-off between the performance and computing cost, we set the number of output channels in the middle layers of DVP to 32. The output of DVP is a spatial attention $P \in [0,1]$ with the size of $H_v\times W_v$, which represents the feature occlusion probability. This process can be modeled as a prediction problem:
\begin{equation}
	P = DVP(F_{vvp}),
	\label{e3}
\end{equation}
where $F_{vvp}$ contains all the intermediate features of VVP. BIFRNet computes the Hadamard product of $F_{c}$ and $P$ to obtain the approximation of visible features $F_v$: 
\begin{equation}
	{F_v}_i = {F_c}_i * P,
\end{equation}
where $F_c$ is the $C_v\times H_v\times W_v$ feature from VVP. $i$ is the channel index. 

\begin{figure}[t]%
	\centering
\includegraphics[width=0.2\textwidth]{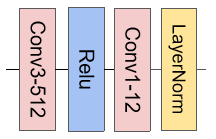}
	\caption{The architecture of the encoder. When training, the encoder receives clean image features and generates the knowledge code $F_k$. When testing, its input is the visible features $F_v$.
	}
	\label{encoder}
\end{figure}

\begin{algorithm}[h]
    \caption{The training of the knowledge matrix $\mathcal K$}
    \label{alg1}
    \begin{algorithmic}[1]
    \REQUIRE ~~\\ 
    The output of the encoder, $F_k$; \\
    The classification label, $l$; \\
    The number of categories, $n$; \\
    \ENSURE  
    $\mathcal K$ \\
   Initialize $\mathcal K$ with a random number matrix; \\
   \WHILE{not done}
   \STATE Initialize $\tilde{F_k}$ with a zero matrix; \\
       \FOR{each $k_i \in F_k$}
       
        \STATE $l_i \Leftarrow$ One-hot$(l[i])$;
        \STATE $\tilde{F_k} \Leftarrow l_i * k_i + \tilde{F_k}$;  
        \ENDFOR
        
        \STATE $//$ b is the batch size
        \STATE $\tilde{\zeta} \Leftarrow \tilde{F_k}/b$; 
        
         \FOR{$j \in n$}
          \IF {$\tilde{\zeta}[j]$ is 0}  
          
          \STATE $\tilde{\zeta}[j] \Leftarrow \mathcal K[j]$;  
          
          \ENDIF
        \ENDFOR
           
       \STATE $L_k \Leftarrow KLD(\tilde{\zeta}, \mathcal K)$;
       
       \STATE minimize $L_k$;
   \ENDWHILE
    \end{algorithmic}
\end{algorithm}

\subsubsection{Prior-Knowledge}
When the occlusion is serious, only visible features of an occluded image cannot provide sufficient information. Thus prior knowledge of objects is required. The proposed knowledge module aims to encode image features to the object knowledge. It contains two components, an encoder and a knowledge matrix $\mathcal K$. Figure~\ref{encoder} shows the architecture of the encoder. 
$\mathcal K$ denotes the knowledge of all categories with size of $N \times H_v \times W_v$, where $N$ denotes the number of categories. 
Each category has a $H_v\times W_v$ feature matrix, which is the knowledge representation. 

$\mathcal K$ is randomly initialized. During training, the encoder encodes features ($\bar{F_c}$ in training and $F_v$ in testing) of images to knowledge coding, called $F_k$. We integrate $F_k$ to form a representation space $\zeta$, which contains knowledge of all categories. It is hoped that $\mathcal K$ is close to $\zeta$. 

Algorithm~\ref{alg1} shows the process of training $\mathcal K$. Each dimension in $\mathcal K$ is expected to represent the knowledge of an category. 
Firstly, $F_k$ of an image is multiplied by the $N$ dimensional one-hot vector converted from its classification label to get $\tilde{F_k}$. 
Values in $\Tilde{F_k}$ are non-zero in the corresponding category dimension and are zero in the non-corresponding ones, which realizes to represent an object. 
Secondly, summing $\tilde{F_k}$ by batch forms a sample of $\zeta$, noted as $\tilde{\zeta}$. If some classes are missing from one batch, the value of the corresponding classification dimensions in $\tilde{\zeta}$ are 0. It should be replaced with the corresponding dimension in $\mathcal K$ to prevent damage to the overall distribution. Finally, by minimizing the Kullback-Leibler divergence (KLD) between $\mathcal K$ and $\tilde{\zeta}$, $\mathcal K$ will represent knowledge of all categories.

After training, knowledge representations in $\mathcal K$ and the similarity matrix of them are shown in Figure~\ref{fusionm}, which demonstrates $\mathcal K$ does have prior knowledge of objects. In the supplementary material, we illustrate the formation process of knowledge. When testing, $\mathcal K$ is input into the completion module as a fixed matrix.

\begin{figure}[t]%
	\centering
	\includegraphics[width=\linewidth]{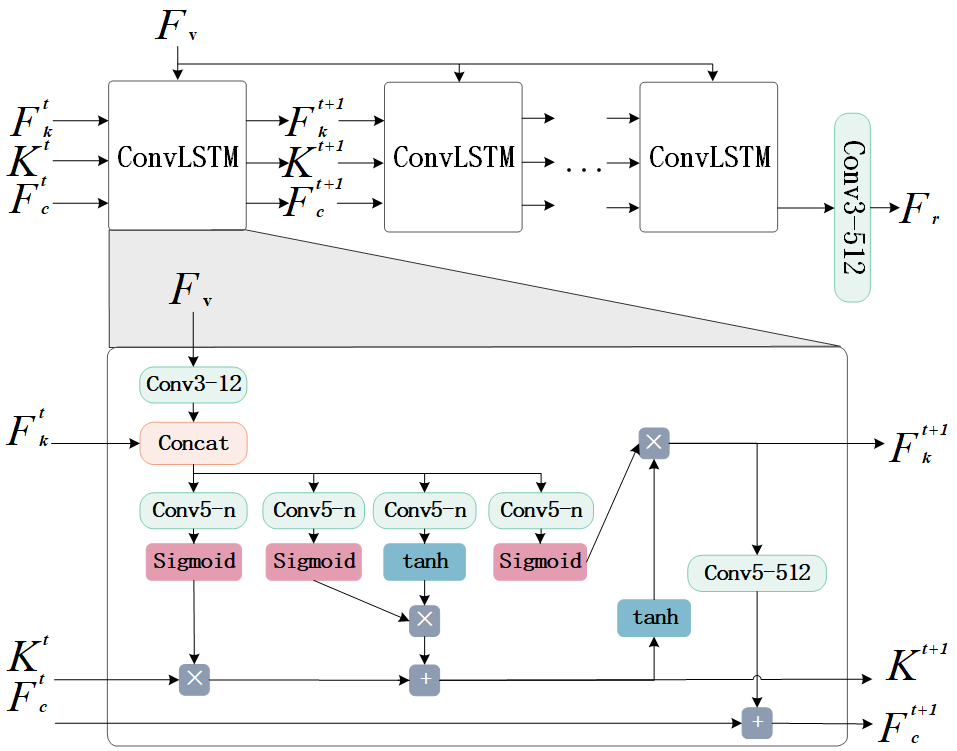}
	\caption{The architecture of the completion module. Its core blocks are ConvLSTM. }
	\label{completion}
\end{figure}

\begin{figure}[t]%
	\centering
	\includegraphics[width=\linewidth]{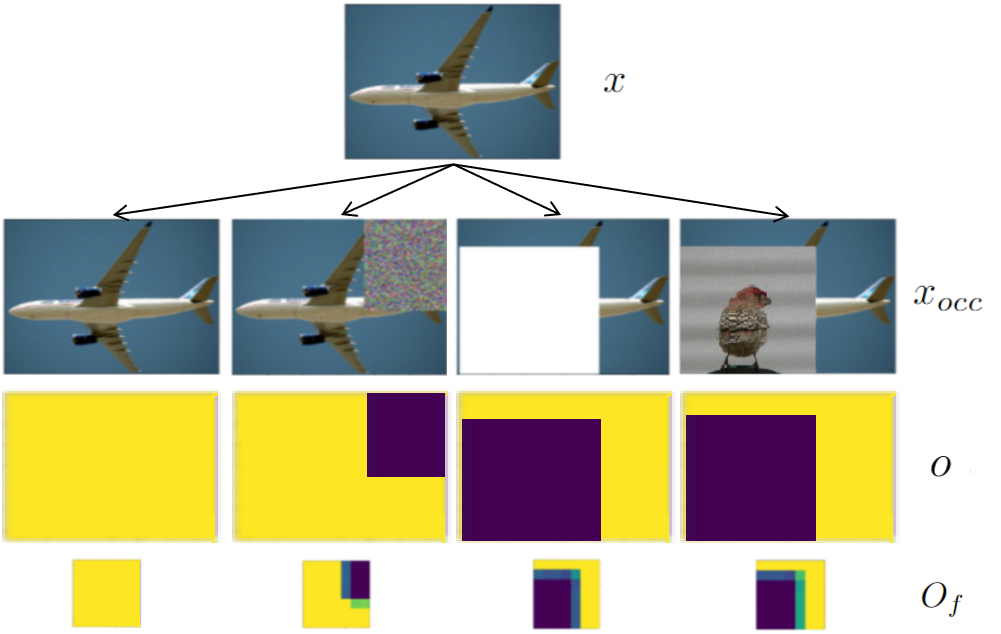}
	\caption{The generation method of the occluded images and its mask labels. Here $x$ is a clean image from the training dataset, and $x_{occ}$ is its occluded version. $o$ and $O_f$ are the occlusion masks of an occluded image and its feature, respectively. Elements in $o$ are 0 when the corresponding postions are occluded and 1 when non-occluded. $O_f$ is obtained by downsampling $o$. There are four occlusion types, unchanged, the noise block, the white block and the image occluders. Here, the image occluders we use are sampled from the first 20 categories in ImageNet valuation set \cite{imagenet} and the KTH-TIPS2-a dataset \cite{KTHTIPS_Databases}. These images are not used to construct the test dataset. The occlusion position and occlusion rate are randomly selected.}
	\label{fig1}
\end{figure}

\subsubsection{Completion}
The completion module restores occluded features, shown as Figure~\ref{completion}. 
The motivation to use the recurrent structure comes from neuroscience \cite{sekuler1992perception,shore1997shape} and the multi-view image generation network GQN \cite{GQN}. 
GQN uses ConvLSTM \cite{convlstm} as the basic block to generate view-specific images based on the representations of other views. 
The completion module in BIFRNet also uses ConvLSTM as the backbone and receives $F_v$, $F_k$, $\mathcal K$ to complete $F_c$, getting the restored feature $F_r$. It can be 
formulated by:
\begin{equation}
\begin{aligned}
	Initial \ state \ \ \ & (F_k^1, \mathcal K^1, F_c^1) = (F_k, \mathcal K, F_c), \\
	State \ update \ \ \ & (F_k^{t+1}, \mathcal K^{t+1}, F_c^{t+1}) \\
	& = ConvLSTM(F_k^t, \mathcal K^t, F_c^{t}, F_v), \\
	Output \ \ \ & F_r = Conv(F_c^7).
\end{aligned}
\end{equation}
The completion module contains 7 ConvLSTM blocks, where $F_c^7$ denotes the features of the last block. 

\begin{table*}[h]
	\centering
		\begin{tabular}{|l|c|clll|llll|llll|}
		\hline
		Occ.Area                         & \multicolumn{1}{l|}{\textbf{L0: 0\%}} & \multicolumn{4}{c|}{\textbf{L1: 20-40\%}}                                                                     & \multicolumn{4}{c|}{\textbf{L2: 40-60\%}}                                                                     & \multicolumn{4}{c|}{\textbf{L3: 60-80\%}}                                                                     \\ \hline
		Occ.Type                         & -                                     & \multicolumn{1}{c|}{w}    & \multicolumn{1}{c|}{n}    & \multicolumn{1}{c|}{t}    & \multicolumn{1}{c|}{o}    & \multicolumn{1}{c|}{w}    & \multicolumn{1}{c|}{n}    & \multicolumn{1}{c|}{t}    & \multicolumn{1}{c|}{o}    & \multicolumn{1}{c|}{w}    & \multicolumn{1}{c|}{n}    & \multicolumn{1}{c|}{t}    & \multicolumn{1}{c|}{o}    \\ \hline
		VGG16         & 99.2                                  & \multicolumn{1}{c|}{96.9} & \multicolumn{1}{l|}{97.0} & \multicolumn{1}{l|}{96.5} & 93.8                      & \multicolumn{1}{l|}{92.0} & \multicolumn{1}{l|}{90.3} & \multicolumn{1}{l|}{89.9} & 79.6                      & \multicolumn{1}{l|}{67.9} & \multicolumn{1}{l|}{62.1} & \multicolumn{1}{l|}{59.5} & 62.2                      \\ \hline
		MaskNet         & 99.6                                  & \multicolumn{1}{c|}{98.6} & \multicolumn{1}{l|}{98.3} & \multicolumn{1}{l|}{98.3} & 96.5                    & \multicolumn{1}{l|}{96.7} & \multicolumn{1}{l|}{96.1} & \multicolumn{1}{l|}{95.3} & 87.6                      & \multicolumn{1}{l|}{84.0} & \multicolumn{1}{l|}{83.1} & \multicolumn{1}{l|}{74.3} & 71.7
		                     \\ \hline
		CoD           & 92.1             & \multicolumn{1}{c|}{92.7} & \multicolumn{1}{c|}{92.3} & \multicolumn{1}{c|}{91.7} & 92.3 & \multicolumn{1}{c|}{87.4} & \multicolumn{1}{c|}{89.5} & \multicolumn{1}{c|}{88.7} & 90.6 & \multicolumn{1}{c|}{70.2} & \multicolumn{1}{c|}{80.3} & \multicolumn{1}{c|}{76.9} & 87.1                   \\ \hline
		VGG+CoD       & 98.3             & \multicolumn{1}{c|}{96.8} & \multicolumn{1}{c|}{95.9} & \multicolumn{1}{c|}{96.2} & 94.4 & \multicolumn{1}{c|}{91.2} & \multicolumn{1}{c|}{91.8} & \multicolumn{1}{c|}{91.3} & 91.4 & \multicolumn{1}{c|}{71.6} & \multicolumn{1}{c|}{80.7} & \multicolumn{1}{c|}{77.3} & 87.2                   \\ \hline
		TDAPNet       & 99.3     & \multicolumn{1}{c|}{98.4} & \multicolumn{1}{c|}{98.9} & \multicolumn{1}{c|}{{98.5}} & 97.4 & \multicolumn{1}{c|}{96.1} & \multicolumn{1}{c|}{97.5} & \multicolumn{1}{c|}{96.6} & 91.6 & \multicolumn{1}{c|}{82.1} & \multicolumn{1}{c|}{88.1} & \multicolumn{1}{c|}{82.7} & 79.8                   \\ \hline
		CompNet-Multi & 99.3             & \multicolumn{1}{c|}{{98.6}} & \multicolumn{1}{c|}{{98.6}} & \multicolumn{1}{c|}{98.8} & {97.9} & \multicolumn{1}{c|}{98.4} & \multicolumn{1}{c|}{98.4} & \multicolumn{1}{c|}{97.8} & {94.6} & \multicolumn{1}{c|}{{91.7}} & \multicolumn{1}{c|}{{90.7}} & \multicolumn{1}{c|}{{86.7}} & {88.4}                   \\ \hline
		BIFRNet    & \textbf{99.8}                                  & \multicolumn{1}{c|}{{\textbf{99.6}}} & \multicolumn{1}{c|}{{\textbf{99.6}}} & 
		\multicolumn{1}{c|}{{\textbf{99.5}}} & \multicolumn{1}{c|}{{\textbf{99.4}}} & \multicolumn{1}{c|}{{\textbf{99.4}}} & \multicolumn{1}{c|}{{\textbf{99.4}}} & \multicolumn{1}{c|}{\textbf{99.2}} & \multicolumn{1}{c|}{{\textbf{98.2}}} & \multicolumn{1}{c|}{\textbf{98.2}} & \multicolumn{1}{c|}{\textbf{98.2}} & \multicolumn{1}{c|}{\textbf{96.2}} & \multicolumn{1}{c|}{{\textbf{95.8}}} \\ \hline
		Human & 100.0                                 & \multicolumn{4}{c|}{100.0}                                                                                    & \multicolumn{4}{c|}{100.0}                                                                                    & \multicolumn{4}{c|}{98.3} \\ \hline
	\end{tabular}
 \caption{Classification results of BIFRNet and some state-of-the-art methods on the Occluded-Vehicles dataset with different levels of artificial occlusion (0\%,20-40\%,40-60\%,60-80\% of the object are occluded) and different types of occluders (w=white boxes, n=noise boxes, t=textured boxes, o=natural objects).}
	\label{t1}
\end{table*}

\subsection{Training Detail}
BIFRNet can be trained end-to-end. Its loss function consists of four parts. 
For DVP, mask labels of occlusion are provided during training. 
We use four kinds of operations to generate occluded images and their corresponding mask labels for each clean image, which is shown in Figure~\ref{fig1}.
Because the convolution operation can maintain the spatial position, we use $O_f$ in Figure\ref{fig1} as the mask label. The size of $O_f$ is $H_v \times W_v$, which is the same size as the output of DVP. 
We use $O_f$ to train DVP to assign high weights to visible parts. Thus the optimization object is to minimize the attention loss, where the Multi-Label Soft Margin Loss (MLSML) is adopted:

\begin{equation}
	\begin{aligned}
	L_{a}(P, O_f) &= - {\frac{1}{N}} \times \sum_{i} O_f[i] \times \log((1 + \exp^{(-P[i])})^{-1}) \\
	&+ (1 - O_f[i]) \times \log(\frac{\exp^{-P[i]}}{1+\exp^{-P[i]}})
	\end{aligned}
\end{equation}


For the knowledge module, we minimize the KLD loss $L_k$, shown in Algrithm~\ref{alg1}.
\begin{equation}
	L_k(\tilde{\zeta} \lVert \mathcal K)  = \sum \tilde{\zeta}(x)\log\frac{\tilde{\zeta}(x)}{\mathcal K(x)}.
\end{equation} 

For the completion module, we suggest that its output $F_r$ should be similar to the features of the corresponding clean image. Therefore, we use the clean image features $\bar{F_c}$ from the pre-trained VGG16 pool5 layer as the target of $F_r$ and minimize the L2 loss of them:
\begin{equation}
	L_{r} = \frac{1}{2} \sum \Vert F_r - \bar{F_c} \Vert^2.
\end{equation}
Finally, cross-entropy loss function is used as the classification loss:
\begin{equation}
	L_{c}(p, y) = \frac{1}{M}\sum{i}\sum_{c=1}^{N}y_{i}^{c}log(p_{i}^{c}).
\end{equation}
where $M$ is the number of samples, $N$ is the number of classes, $y_i^c$ is the label, and $p_i^c$ is the predicted classification probability of the model. Totally, the loss function is as follows:
\begin{equation}
	L = L_k + L_{r} + L_{c} + L_{a} \times \alpha.
\end{equation}
Here, $\alpha$ is a weight parameter, and we set it 0.1. Discussion about it is presented in the supplementary material. After training, given an input image, BIFRNet will output the prediction result.

\section{Experiments} \label{experiments}
\subsection{Dataset}
We use Occluded-Vehicles dataset and Occluded-COCO Vehicles dataset \cite{Kortylewski_2020_WACV} to evaluate the proposed BIFRNet. For Occluded-Vehicles dataset, images are from the PASCAL3D+ dataset \cite{PASCAL3D+}, and the occluded images are generated artificially by covering four different types of occluders (\emph{i.e.}, white, noise, textures, and segmented objects) to the clean images. Four occlusion levels are used to represent the percentage of occluded area: Level-0 (0\%), Level-1 (20-40\%), Level-2 (40-60\%), and Level-3 (60-80\%). Images in Occluded-COCO Vehicles dataset are selected from MS-COCO \cite{mscoco} dataset and also has four occlusion levels. Especially, occluded images in this dataset are real-world. The two datasets both contain 12 different categories. Following \cite{topdown}, we use all categories for training and use 6 categories including aeroplane, bicycle, bus, car, motorbike, and train, for testing. Detailed descriptions can be found in the supplementary material, accessible at https://github.com/JiaHongZ/BiFRNet.

\begin{table}[t]
	\centering
		\begin{tabular}{|l|cccc|}
		\hline
		
		Occ.Area        & \multicolumn{1}{c|}{\textbf{L0}} & \multicolumn{1}{c|}{\textbf{L1}} & \multicolumn{1}{c|}{\textbf{L2}} & \textbf{L3} \\ \hline
		VGG16           & \multicolumn{1}{c|}{99.1}        & \multicolumn{1}{c|}{88.7}        & \multicolumn{1}{c|}{78.8}        & 63.0        \\
		MaskNet           & \multicolumn{1}{c|}{\textbf{99.5}}        & \multicolumn{1}{c|}{88.5}        & \multicolumn{1}{c|}{78.8}        & 65.8        \\ 
		TDAPNet        & \multicolumn{1}{c|}{99.4}        & \multicolumn{1}{c|}{88.8}        & \multicolumn{1}{c|}{87.9}        & 69.9        \\  
		CompNet-Multi     & \multicolumn{1}{c|}{99.4}        & \multicolumn{1}{c|}{{\textbf{95.3}}}        & \multicolumn{1}{c|}{90.9}        & \textbf{86.3}        \\ \hline
		BIFRNet         & \multicolumn{1}{c|}{\textbf{99.5}}        & \multicolumn{1}{c|}{94.8}        & \multicolumn{1}{c|}{\textbf{93.5}}        & \textbf{86.3}        \\ \hline
	\end{tabular}
 \caption{Classification results on the Occluded-COCO Vehicles dataset.}
	\label{t2}
\end{table}

\subsection{Experimental Settings}
We use Adam \cite{adam} as the optimizer. 
Batch size and the initial learning rate are set to 64 and 0.0001, respectively, and the learning rate is cosine decayed. 
VGG16 used in this paper is pre-trained in ImageNet \cite{imagenet}. The training and testing settings for Occluded-Vehicles and Occluded-COCO Vehicles datasets stay the same. 

\begin{table*}[t]\small
	\centering
	\begin{tabular}{|p{2.1cm}|c|clll|llll|llll|}
		\hline
		Occ.Area   & \multicolumn{1}{l|}{\textbf{L0: 0\%}} & \multicolumn{4}{c|}{\textbf{L1: 20-40\%}}                                                                                                                          & \multicolumn{4}{c|}{\textbf{L2: 40-60\%}}                                                                                                                         & \multicolumn{4}{c|}{\textbf{L3: 60-80\%}}                                                                                                                          \\ \hline
		Occ.Type   & -                                     & \multicolumn{1}{c|}{w}             & \multicolumn{1}{c|}{n}             & \multicolumn{1}{c|}{t}             & \multicolumn{1}{c|}{o}                   & \multicolumn{1}{c|}{w}             & \multicolumn{1}{c|}{n}             & \multicolumn{1}{c|}{t}             & \multicolumn{1}{c|}{o}              & \multicolumn{1}{c|}{w}             & \multicolumn{1}{c|}{n}             & \multicolumn{1}{c|}{t}             & \multicolumn{1}{c|}{o}                     \\ \hline
		$\mathcal K+N(0,1)$     & \textbf{99.84}                                             & \multicolumn{1}{l|}{\textbf{99.61}}          & \multicolumn{1}{l|}{99.61}          & \multicolumn{1}{l|}{\textbf{99.52}}            & \multicolumn{1}{l|}{\textbf{99.40}}          & \multicolumn{1}{l|}{99.42}          & \multicolumn{1}{l|}{\textbf{99.38}}          & \multicolumn{1}{l|}{\textbf{99.22}}                 & \multicolumn{1}{l|}{98.20}          &
		\multicolumn{1}{c|}{98.23}         &
		\multicolumn{1}{l|}{98.20}          & \multicolumn{1}{l|}{96.11}          & \multicolumn{1}{l|}{\textbf{95.90}}                   \\ \hline
		$\mathcal K+N(0,3)$     & 99.82                                             & \multicolumn{1}{l|}{\textbf{99.61}}          & \multicolumn{1}{l|}{99.61}          & \multicolumn{1}{l|}{99.49}            & \multicolumn{1}{l|}{99.29}          & \multicolumn{1}{l|}{\textbf{99.45}}          & \multicolumn{1}{l|}{\textbf{99.38}}          & \multicolumn{1}{l|}{99.17}                 & \multicolumn{1}{l|}{98.16}          &
		\multicolumn{1}{c|}{\textbf{98.25}}         &
		\multicolumn{1}{l|}{98.18}          & \multicolumn{1}{l|}{95.92}          & \multicolumn{1}{l|}{95.62}                   \\ \hline
		$\mathcal K+N(0,5)$     & 99.82                                            & \multicolumn{1}{l|}{\textbf{99.61}}          & \multicolumn{1}{l|}{99.61}          & \multicolumn{1}{l|}{99.49}            & \multicolumn{1}{l|}{99.29}          & \multicolumn{1}{l|}{\textbf{99.45}}          & \multicolumn{1}{l|}{\textbf{99.38}}          & \multicolumn{1}{l|}{99.17}                 & \multicolumn{1}{l|}{97.93}          &
		\multicolumn{1}{c|}{98.23}         &
		\multicolumn{1}{l|}{98.13}          & \multicolumn{1}{l|}{95.85}          & \multicolumn{1}{l|}{94.98}                   \\ \hline
		BIFRNet-$K$        &	99.72                                  & \multicolumn{1}{l|}{99.54}          & \multicolumn{1}{l|}{\textbf{99.65}}          & \multicolumn{1}{l|}{99.38}            & \multicolumn{1}{l|}{99.22}          & \multicolumn{1}{l|}{99.24}          & \multicolumn{1}{l|}{99.29}          & \multicolumn{1}{l|}{98.76}                 & \multicolumn{1}{l|}{97.86}          &
		\multicolumn{1}{c|}{97.51}         &
		\multicolumn{1}{l|}{97.77}          & \multicolumn{1}{l|}{94.15}          & \multicolumn{1}{l|}{93.73}           \\ \hline
		BIFRNet    & \textbf{99.84}                                  & \multicolumn{1}{c|}{{\textbf{99.61}}} & \multicolumn{1}{c|}{{99.61}} & 
		\multicolumn{1}{c|}{{\textbf{99.52}}} & \multicolumn{1}{c|}{{\textbf{99.40}}} & \multicolumn{1}{c|}{{99.42}} & \multicolumn{1}{c|}{{\textbf{99.38}}} & \multicolumn{1}{c|}{\textbf{99.22}} & \multicolumn{1}{c|}{{\textbf{98.23}}} & \multicolumn{1}{c|}{98.18} & \multicolumn{1}{c|}{\textbf{98.25}} & \multicolumn{1}{c|}{\textbf{96.15}} & \multicolumn{1}{c|}{{95.81}} \\ \hline
	\end{tabular}
 	\caption{Studies of Knowledge in BIFRNet on the Occluded-Vehicles dataset.}
	\label{tknowledge}
\end{table*}

\begin{table*}[t]
	\centering
	\begin{tabular}{|l|c|clll|llll|llll|}
		\hline
		Occ.Area   & \multicolumn{1}{l|}{\textbf{L0: 0\%}} & \multicolumn{4}{c|}{\textbf{L1: 20-40\%}}                                                                                                                          & \multicolumn{4}{c|}{\textbf{L2: 40-60\%}}                                                               & \multicolumn{4}{c|}{\textbf{L3: 60-80\%}}                                                                \\ \hline
		Occ.Type   & -                                     & \multicolumn{1}{c|}{w}             & \multicolumn{1}{c|}{n}             & \multicolumn{1}{c|}{t}             & \multicolumn{1}{c|}{o}                   & \multicolumn{1}{c|}{w}             & \multicolumn{1}{c|}{n}             & \multicolumn{1}{c|}{t}             & \multicolumn{1}{c|}{o}              & \multicolumn{1}{c|}{w}             & \multicolumn{1}{c|}{n}             & \multicolumn{1}{c|}{t}             & \multicolumn{1}{c|}{o}                     \\ \hline
		Completion Ablation     & 99.4                                            & \multicolumn{1}{l|}{98.8}          & \multicolumn{1}{l|}{99.0}          & \multicolumn{1}{l|}{98.9}            & \multicolumn{1}{l|}{98.5}          & \multicolumn{1}{l|}{98.5}          & \multicolumn{1}{l|}{98.6}          & \multicolumn{1}{l|}{98.4}                 & \multicolumn{1}{l|}{97.1}          &
		\multicolumn{1}{c|}{96.1}         &
		\multicolumn{1}{l|}{96.1}          & \multicolumn{1}{l|}{92.9}         & \multicolumn{1}{l|}{92.6}                   \\ \hline
		BIFRNet-Completion        &	99.7                                 & \multicolumn{1}{c|}{99.4} & \multicolumn{1}{l|}{99.3} & \multicolumn{1}{l|}{{99.3}} & 99.1          & \multicolumn{1}{l|}{99.1} & \multicolumn{1}{l|}{98.9} & \multicolumn{1}{l|}{{98.7}} & 98.0      & \multicolumn{1}{l|}{{97.3}} & \multicolumn{1}{l|}{96.8} & \multicolumn{1}{l|}{93.5} & 92.8       \\ \hline
		BIFRNet    & \textbf{99.8}                                  & \multicolumn{1}{c|}{{\textbf{99.6}}} & \multicolumn{1}{c|}{{\textbf{99.6}}} & 
		\multicolumn{1}{c|}{{\textbf{99.5}}} & \multicolumn{1}{c|}{{\textbf{99.4}}} & \multicolumn{1}{c|}{{\textbf{99.4}}} & \multicolumn{1}{c|}{{\textbf{99.4}}} & \multicolumn{1}{c|}{\textbf{99.2}} & \multicolumn{1}{c|}{{\textbf{98.2}}} & \multicolumn{1}{c|}{\textbf{98.2}} & \multicolumn{1}{c|}{\textbf{98.2}} & \multicolumn{1}{c|}{\textbf{96.2}} & \multicolumn{1}{c|}{{\textbf{95.8}}} \\ \hline
	\end{tabular}
 	\caption{Studies of Completion in BIFRNet on the Occluded-Vehicles dataset.}
	\label{tcompletion}
\end{table*}

\subsection{Experimental Results} \label{er}

We compare BIFRNet to VGG16 \cite{VGG16} and some state-of-the-art models CoD \cite{kortylewski2020combining}, TDAPNet \cite{topdown} and CompNet-Mul \cite{Kortylewski_2020_WACV} on Occluded-Vehicles and Occluded-COCO Vehicles. MaskNet \cite{maskatt} was proposed to recognize human face and its backbone is ResNet \cite{resnet}. Here we re-implement it using VGG16 for a fair comparison. 
The human experimental data is from \cite{kortylewski2020combining}.

\textbf{Results on Occluded-Vehicles}: Table~\ref{t1} shows the comparison results on the synthetic dataset Occluded-Vehicles. We observe that BIFRNet significantly outperforms other methods at all occlusion levels, and achieves human-like performances. 
As for occlusion types, BIFRNet improves the classification accuracy on the object occlusion most. 

\textbf{Results on Occluded-COCO Vehicles}: 
Here we evaluate BIFRNet on the real-world Occluded-COCO Vehicles dataset. Experimental results reported in Table~\ref{t2} show that BIFRNet still outperforms the other methods on real-world occluded images.

\begin{figure}[]%
	\centering
	\includegraphics[width=\linewidth]{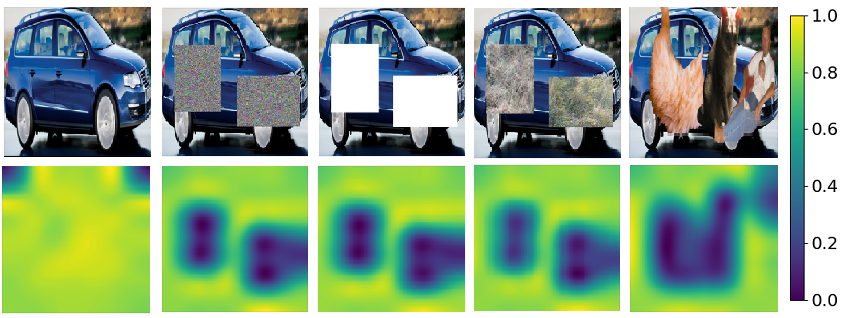}
	\caption{Visualization of the attention maps $P$ from the dorsal visual pathway, where the bright parts have higher attention than the darker ones. The upper row includes images from the Occluded-Vehicles dataset, and the lower row shows the corresponding attention map $P$.}
	\label{att1}
\end{figure}

\section{Discussion} \label{ablation}
This section discusses how BIFRNet's three main components (visual pathways, the knowledge module, and the completion module) affect the performance.

\noindent\textbf{DVP}. \quad
DVP in BIFRNet generates the spatial attention, which distinguishes occluders from target objects. We visualize the attention map $P$ in Figure~\ref{att1}. The actual size of the attention map is $7\times7$, and we upsample them to the same size of input images for convenient observation. 
Figure~\ref{att1} visualizes a car image from the testing set of Occluded-Vehicles under different occluders at L2 occlusion level as well as the corresponding attention map.
The bright regions in the attention map have higher weights than the darker regions, which indicates that DVP learns to discriminate occluded and non-occluded features.
Moreover, DVP can be generalized to tackle real-world occlusion. Detailed study can be found in the supplementary material.


\noindent\textbf{Knowledge}. \quad
Here we conduct two experiments to evaluate the knowledge module. 
In the first experiment, we perturb $\mathcal K$ by a random matrix sampled from the normal distribution $N(\mu, \sigma)$. 
We denote the first experiment as $\mathcal K + N(\mu, \sigma)$ and report the results in Table~\ref{tknowledge}. 
It is observed that the recognition accuracy gradually decreases as the random perturbation increases, which indicates that the knowledge affects BIFRNet's recognition results. 
In the second experiment, we remove the knowledge module and retrain BIFRNet.
This experiment is denoted as BIFRNet-$K$ and the corresponding results in Table~\ref{tknowledge} demonstrate that the knowledge module is important for BIFRNet to achieve advanced performances, especially at high occlusion level (L3). Furthermore, we observe that changing $\mathcal K$ will achieve top-down knowledge modulated recognition , which is discussed in the supplementary material.

\noindent\textbf{Completion}. \quad
We studied the completion module from two aspects. At first, we tested BIFRNet by cutting off the completion module, that is, using $F_v$ as the classification feature. The results are shown in Table~\ref{tcompletion}, noted as Completion Ablation. It can be found that BIFRNet's classification accuracy decreases, and the more serious the occlusion exists, the more the accuracy decreases. BIFRNet-Completion in Table~\ref{tcompletion} denotes removing the completion module and retraining BIFRNet. Results show that the recognition performance reduces significantly when serious occlusion occurs. It indicates that the restored features is more robust for occlusion than the visible features. 

\section{Conclusion} \label{conclusion}
This paper proposes a novel brain-inspired feature restoration network BIFRNet. It achieves higher performance than the previous methods for POIR problem on both the synthetic and real-world occluded image datasets. All the components of BIFRNet make contributions to this superior performance. The visual pathways provide BIFRNet with image features, the knowledge module provides prior object knowledge, and the completion module restores the occluded features. Although we use occluded images to train BIFRNet, experiments show that BIFRNet is also robust to the occluders it has never seen. BIFRNet can be trained and tested end-to-end, which is convenient for actual applications. We state that a robust occlusion framework should have the abilities of occlusion discrimination, prior knowledge, and feature restoration. This brain-inspired framework may provide instructive ideas for the future research on POIR and other POIR-like problems. 

\section{Acknowledgments}
This paper is supported by supported by the STI 2030—Major Projects (grant No. 2021ZD0200300) and the National Natural Science Foundation of China (grant No. 62176241).

\bibliography{bibfile}

\end{document}